\documentclass{article}

\usepackage[nonatbib,preprint]{neurips_2024}

\usepackage[utf8]{inputenc} % allow utf-8 input
\usepackage[T1]{fontenc}    % use 8-bit T1 fonts
\usepackage{hyperref}       % hyperlinks
\usepackage{url}            % simple URL typesetting
\usepackage{booktabs}       % professional-quality tables
\usepackage{amsfonts}       % blackboard math symbols
\usepackage{nicefrac}       % compact symbols for 1/2, etc.
\usepackage{microtype}      % microtypography
\usepackage{xcolor}         % colors
\usepackage{adjustbox}
\usepackage{amsmath}
\usepackage{color}
\usepackage{algorithm}
\usepackage{algorithmic}
\usepackage{setspace}
\usepackage{wrapfig}
\usepackage{enumitem}
\usepackage{pifont} % for checkmark and cross
\setlist{topsep=0pt, leftmargin=*}
\usepackage{graphicx}
\usepackage{caption}
\usepackage{subcaption}
\usepackage{multirow}
\usepackage{makecell}

\usepackage[style=numeric,sorting=none]{biblatex}
\newcommand{\Name}{Uni-Mol2\space}

\title{Uni-Mol2: Exploring Molecular Pretraining Model at Scale}

\author{% 
Xiaohong Ji$^{1}$\thanks{\scriptsize Equal contribution.}, Zhen Wang$^{1}$\footnotemark[1], Zhifeng Gao$^1$\footnotemark[1], Hang Zheng$^1$  \\
\textbf{Linfeng Zhang$^{1,2}$,} \textbf{Guolin Ke$^1$, Weinan E$^{2,3,4}$}  \\
\quad $^1$DP Technology, Beijing, 100080, China. \\ 
\quad $^2$AI for Science Institute, Beijing 100080, China. \\ 
\quad $^3$School of Mathematical Sciences, Peking University, Beijing, 100871, China. \\
\quad $^4$Center for Machine Learning Research, Peking University, Beijing 100084, China. \\
\texttt{\{jixh, wangz, gaozf, zhengh, zhanglf, kegl\}@dp.tech} \\
\texttt{weinan@math.pku.edu.cn}
}

\begin{document}

\maketitle
\begin{abstract}
In recent years, pretraining models have made significant advancements in the fields of natural language processing (NLP), computer vision (CV), and life sciences. The significant advancements in NLP and CV are predominantly driven by the expansion of model parameters and data size, a phenomenon now recognized as the scaling laws. However, research exploring scaling law in molecular pretraining models remains unexplored. In this work, we present \Name, an innovative molecular pretraining model that leverages a two-track transformer to effectively integrate features at the atomic level, graph level, and geometry structure level. Along with this, we systematically investigate the scaling law within molecular pretraining models, characterizing the power-law correlations between validation loss and model size, dataset size, and computational resources. Consequently, we successfully scale \Name to 1.1 billion parameters through pretraining on 800 million conformations, making it the largest molecular pretraining model to date. Extensive experiments show consistent improvement in the downstream tasks as the model size grows. The \Name with 1.1B parameters also outperforms existing methods, achieving an average 27\% improvement on the QM9 and 14\% on COMPAS-1D dataset.
\end{abstract}

\section{Introduction}

With the exponential growth of available biological data, there arises a critical need for innovative computational methodologies to utilize this wealth of information effectively. While traditional molecular representations like fingerprint-based models \cite{yang2019machine-fp1, rogers2010extended-ecfp} lack the ability to capture fine-grained structural features and struggle to handle large or complex molecules effectively. Molecular Representation Learning (MRL) using molecular pretraining emerges as a promising approach, leveraging the power of machine learning to imbue algorithms with a deep understanding of molecular structures and functions. Various modalities of molecular representation by pretraining have been extensively studied in the past. The typical approach for representing molecules involves two main strategies. One strategy is to represent molecules as one-dimensional sequential strings, such as SMILES \cite{wang2019smiles-bert, xu2017seq2seq} and InChI \cite{winter2019learning-ich1}. The representative work is SMILES-BERT\cite{wang2019smiles-bert}, which learns from large-scale unlabeled data through the masked SMILES recovery task. Another strategy is to represent molecules as two-dimensional graphs \cite{fang2022geometry-gem, rong2020self-grover, wang2022molecular-molcl}. MolCLR \cite{wang2022molecular-molcl}, a typical method, learns the representations from unlabeled data by contrasting positive molecule graph pairs against negative ones. Additionally, a growing trend is to leverage three-dimensional information in MRL to enable tasks like 3D geometry prediction or generation \cite{stark20223d, liu2021pre, zhou2023unimol}. The pursuit of molecular pretraining has sparked a wave of exploration and innovation across the field, marking a new era of discovery within the discipline.

While in the past few years, scaling up pre-trained language models \cite{radford2019language-gpt2, brown2020language-gpt3, achiam2023gpt4, deepseek-llm, touvron2023llama, liu2024visual, chen2023internvl} has been achieved remarkable progress in natural language processing (NLP) and computer vision (CV). The exponential growth in model size and the richness of training data have significantly enhanced the capabilities and performance of LLMs across various NLP and CV tasks. Despite extensive research on molecular pretraining, the majority of prior studies have been conducted on a relatively small scale, utilizing limited parameters and datasets. Learning scalable molecular representation learning is rarely explored and remains a challenging problem. The recent \cite{chen2024uncovering}'s work conducts a series of data-centric experiments to demonstrate scaling behaviors in various aspects. The exploration of the molecular pretraining model is limited to the GIN \cite{xu2018powerful-gin}, SchNet \cite{schutt2017schnet}, whose model scale and data scale are comparatively small.

To delve deeper into the scaling of molecular pretraining foundational models, our preliminary investigations have yielded notable insights within this domain. We summarize the contributions of this work as follows:

\begin{itemize}

\item We have curated and organized a dataset comprising approximately 884 million 3D conformations, which contains 73 million scaffolds for pretraining. To the best of our knowledge, this is the largest dataset of molecules with 3D conformations for molecular pretraining to date, which provides the foundation ingredient for training large-scale molecular models.

\item We systematically study the scalability and flexibility of Uni-Mol2 in terms of model parameters, which range from 84M to 1.1B parameters, and characterize the relationship between validation loss and model size, dataset size, and computational resources. It is the first time to demonstrate the scaling law of molecular pretraining and Uni-Mol2 is currently the largest billion-scale molecular pretraining model to date.

\item We present an in-depth analysis of scaling trends about fine-tuning on downstream tasks as the results are shown in Table\ref{tab:qm9 performance} and \ref{tab:photoelectric performance}, Uni-Mol2 demonstrates consistent improvement in downstream task performance with increasing model parameters. The 1.1 billion parameters model also achieves significant improvement over the existing method.

\end{itemize}

\section{Related Work}

\paragraph{Molecular representation learning} Previous research has extensively investigated various modalities for molecular representation. A range of methods have been proposed based on different types of information utilized during pretraining. SMILES-BERT\cite{wang2019smiles-bert} uses the smiles sequence in pretraining to capture the representation. Due to SMILES representation lack of explicit encoding of molecular structural information. To address this limitation, GROVER integrates Message Passing Networks into a Transformer-style architecture and learns from unlabeled molecular data through carefully designed self-supervised tasks at different levels of molecular topology (node, edge, and graph). Furthermore, GEM\cite{fang2022geometry-gem} incorporates three-dimensional (3D) spatial structure information, atoms, bonds, and bond angles simultaneously to model the molecular representation.
\vspace{-10pt}
\paragraph{Foundation models}
Recently, there has been considerable interest in developing foundational models to consolidate and expand representations. The significant advancements in scaling up pre-trained language models\cite{radford2019language-gpt2, brown2020language-gpt3, achiam2023gpt4} have fundamentally reshaped the field of natural language processing. \cite{bai2023qwen, deepseek-llm, chen2023internvl, jiang2023mistral} also prove that the foundation model demonstrates strong performance on many NLP datasets, sometimes reaching or exceeding the human performance. Some works in CV\cite{zhai2022scaling-svt, dehghani2023scaling-svt22b} demonstrate the potential for “LLM-like” scaling in vision and 
underscore significant improvement via model and data scaling. And Sora\cite{sora, liu2024sora-survey}, a multi-modal foundation model exhibits the capacity to offer sophisticated understanding regarding the intricate interplay of physical and contextual dynamics within depicted scenes.

\section{Pretraining}

The pretraining stage of molecular involves learning from vast amounts of molecular data to acquire a comprehensive understanding of molecular representations. By pretraining on a large and diverse unlabeled dataset, the model can develop a rich understanding of molecular structures and properties, which can subsequently be fine-tuned or applied to specific downstream tasks, such as drug discovery, materials design, or chemical synthesis. The section provides details of the data curation process for pretraining, the detailed pretraining architecture, the well-designed self-supervision tasks, and the specific training procedures employed for scaling up the model.

\subsection{Data}

\begin{figure*}[t]
    \centering
    \includegraphics[width=\textwidth]{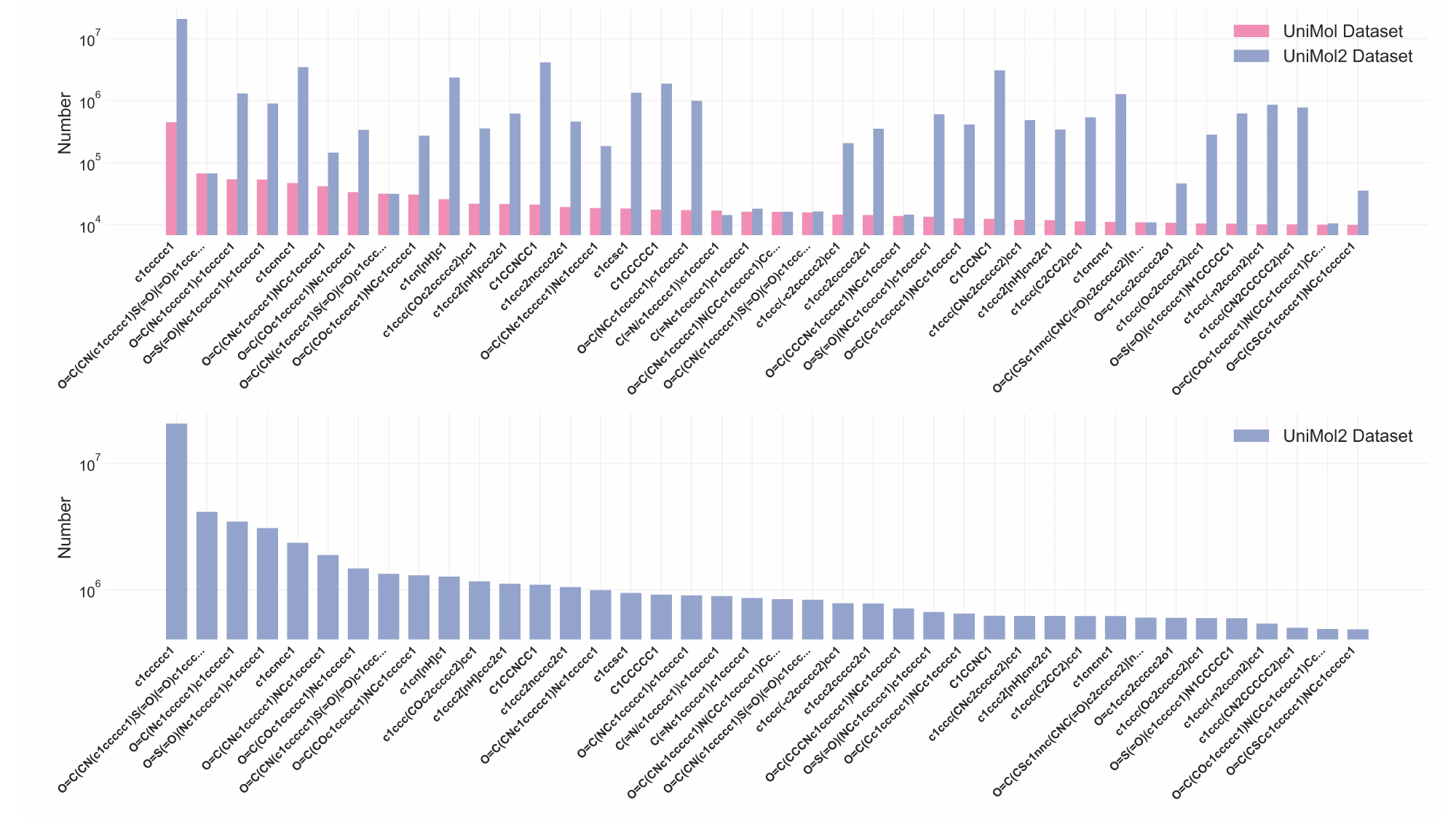}
    \caption{\textbf{Top}: Comparison of scaffold frequency between Uni-Mol and Uni-Mol2 dataset.  \textbf{Bottom}: Scaffolds distribution on Uni-Mol2 dataset}
    \label{fig:scaffold frequency comparison Uni-Mol1 and Uni-Mol2 dataset}
\end{figure*}

To augment the richness and diversity of the dataset, we integrated the two parts we have collected. One part consists of approximately 19 million molecules sourced from Uni-Mol \cite{zhou2023unimol}, while the other is derived from ZINC20 \cite{irwin2020zinc20,} which includes 1.4 billion compounds. We downloaded the subset with standard reactivity, which contains 884 million compounds from website \footnote{https://zinc20.docking.org/tranches/home/}. Table \ref{tab:v2 dataset} shows the enrichment compared with Uni-Mol dataset. The overall Uni-Mol2 dataset has increased by over 40 times compared to the Uni-Mol dataset, with the number of scaffold increasing by 17 times, greatly expanding the diversity of the data. Figure \ref{fig:scaffold frequency comparison Uni-Mol1 and Uni-Mol2 dataset}(\textbf{Top}) shows the numeric distributions of the top 40 skeletons in Uni-Mol dataset and the number corresponding in Uni-Mol2 dataset. To prevent data leakage in evaluating pretraining performance, we randomly sampled 520k molecules from the Uni-Mol2 dataset as the validation set to evaluate the effectiveness and investigate the scaling relationship.

As illustrated in the visualization depicting the frequency distribution of the top 40 Murcko scaffolds in Uni-Mol2 dataset (refer to Figure \ref{fig:scaffold frequency comparison Uni-Mol1 and Uni-Mol2 dataset} (\textbf{Bottom})), it is observed that the molecular scaffold conforms to a distribution characterized by a long-tail pattern. To create a more balanced training dataset, we categorize the SMILES of Uni-Mol2 training set by Murcko scaffold, resulting in 73,725,454 scaffolds along with frequency distribution. Then, We utilize the temperature-based sampling method \cite{conneau2019cross}\cite{chung2023unimax}, as described in equation \ref{sample equation} to select molecules from Uni-Mol2 training set. 

\begin{equation}
\begin{aligned}
\label{sample equation}
P_{i} &= \frac{N_{s_{i}}}{\sum{N_{s_{i}}}}, \\
P_{scaffold_{i}} &= \text{softmax}(\frac{P_{i}}{\tau})
\end{aligned}
\end{equation}

Where $N_{s{i}}$ represents the number of molecules with $i$-th scaffold in Uni-Mol2 training set. The temperature $\tau$ modulates the smoothness of the molecular distribution across scaffolds. We use an empirical value $\tau = 0.005 $ as the temperature to effectively balance the proportion of molecules with high-frequency and low-frequency scaffolds.

\begin{table*}
\caption{The different scale of Uni-Mol dataset and Uni-Mol2 dataset}
% \vspace{8pt}
\centering
\begin{tabular}{c|c|c|c}
    \toprule
        Datasets     & SMILES     & Scaffold & Data Source  \\
        \midrule
        Uni-Mol Dataset &  19M & 4,224,621 & ZINC15, ChemBL, Commercial\ Database\cite{zhou2023unimol} \\ 
        Uni-Mol2 Dataset &  $\sim$ 884M & 73,725,454 & Uni-Mol Dataset, ZINC 20\cite{irwin2020zinc20} \\
    \bottomrule
\end{tabular}
\label{tab:v2 dataset}
\end{table*}

\subsection{Architecture}

\graphicspath{{}}
\begin{figure*}[t]
    \centering
    \includegraphics[width=\textwidth]{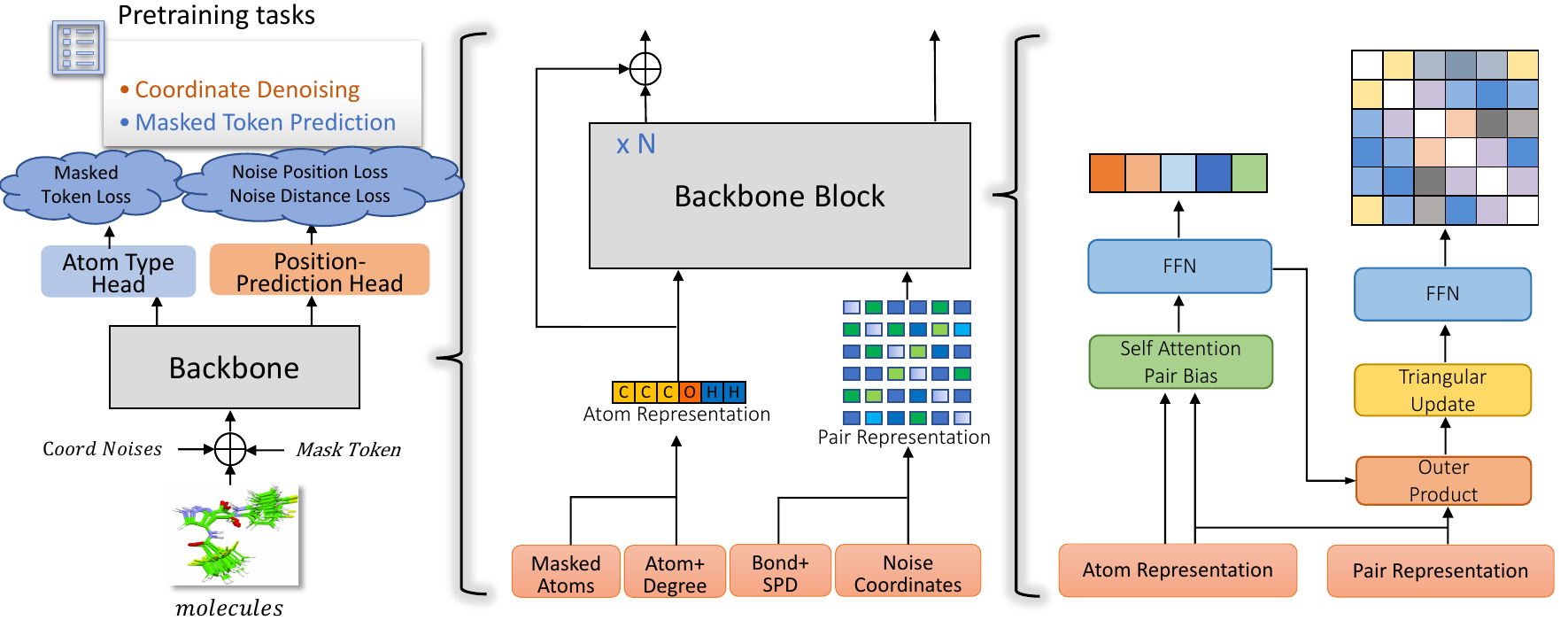}
    \caption{\textbf{Left}: The overall pretraining architecture. \textbf{Middle}: Atom and Pair representation. \textbf{Right}: The details of backbone block}
    \label{fig:Uni-Mol2 architecture}
\end{figure*}

As depicted in Figure \ref{fig:Uni-Mol2 architecture}, Uni-Mol2 essentially adheres to the model design of Uni-Mol+\cite{lu2023highly}, acting as a two-track transformer that concurrently processes atom features and pair features. Consistent with Uni-Mol\cite{zhou2023unimol}, Uni-Mol2 employs two self-supervised tasks: masked token prediction and molecule coordinate denoising. The detailed framework is presented as follows:

\textbf{Feature Representation and Position Encoding} Given molecular $M = (x, e, r)$,  where $x \in \mathbb{R}^{n \times d_{a}}$ denotes atom features, $e \in \mathbb{R}^{n \times n \times d_{e}}$ denotes bond features and $r \in \mathbb{R}^{n \times 3}$ denotes coordinate features. Following Uni-Mol+, we employ RDKit to obtain atom token $x_{\text{token}}^{i}$, atom degree $x_{\text{degree}}^{i}$, and atomic features $x_{\text{atomic}}^{i}$ for each atom. The atom embedding $x_{\text{atom}}^{i}$ is then initialized as:
\begin{equation}
\label{eq:atom emb}
x_{\text{atom}}^{i} = \text{Embedding}(x_{\text{token}}^{i}) + \text{Embedding}(x_{\text{degree}}^{i}) + \text{Embedding}(x_{\text{atomic}}^{i})
\end{equation}

For pair features, we utilize RDKit to obtain bond features $x_{\text{bond}}^{i,j}$ by $\text{Embedding}(x_{\text{bond}}^{i,j})$. We adopt the method from \cite{ying2021transformers, lu2023highly} to encode the shortest path distance $x_{\text{SPD}}^{i,j}$ of atom pair (i, j) in the molecular graph by $\text{Embedding}(x_{\text{SPD}}^{i,j})$. Additionally, we employ the Gaussian kernel approach with pair type, as described in \cite{shi2022benchmarking, zhou2023unimol}, to encode the Euclidean distance of the atom pair (i, j) by $\psi^{i,j}$.  
The pair embedding $x_{\text{pair}}^{i,j}$ is then initialized as:

\begin{equation}
\label{eq:pair emb}
x_{\text{pair}}^{i,j} = \text{Embedding}(x_{\text{bond}}^{i,j}) + \text{Embedding}(x_{\text{SPD}}^{i,j}) + \psi^{i,j}
\end{equation} 

\textbf{Two-track Transformer Layer}
The backbone of Uni-Mol2 has $N$ blocks, each block handles atom representation and pair representation concurrently. Formally, for the $l$-th block, Uni-Mol2 update atom representation $x^{l}$ by 

\begin{equation}
\begin{aligned}
\label{eq:atom attention}
x^{l} &= \text{SelfAttentionPairBias}(\text{LN}(x^{l-1}), p^{l-1}), \\
x^{l} &= x^{l-1} + \text{FFN}(\text{LN}(x^{l}))
\end{aligned}
\end{equation}

For the pair representation $p^{l}$, 
\begin{equation}
\begin{aligned}
\label{eq: pair representation update}
p^{l} &= p^{l-1} + \text{OuterProduct}(\text{LN}(p^{l-1})),\\
p^{l} &= p^{l} + \text{TriangularUpdate}(\text{LN}(p^{l})),\\
p^{l} &= p^{l} + \text{FFN}(\text{LN}(p^{l}))
\end{aligned}
\end{equation}

The details of SelfAttentionPairBias, OuterProduct, and TriangularUpdate are aligned with those of Uni-Mol+. Additionally, Uni-Mol2 adopts pre-norm layer normalization at atom and pair representation, which differs from Uni-Mol+, to improve stability in the model's training dynamics. Specifically, we set atom embedding $x_{\text{atom}}$ as atom representation $x^{0}$ and pair embedding $x_{\text{pair}}$ as pair representation $p^{0}$ for the first block.

\textbf{Pretraining Tasks}
To effectively model the structure of molecular conformations, we set pretraining tasks basically following Uni-Mol. In detail, for each molecule, we randomly mask 15\% of the atom tokens with the placeholder token $[\text{MASK}]$. We then add the atom token prediction head to optimize masked atom token loss $\mathcal{L}_{\text{atom}}$ by 
\begin{equation}
\label{eq:ce loss}
\mathcal{L}_{\text{atom}} = H(x_{\text{atom}}[\text{mask}], x_{\text{patom}}[\text{mask}])
\end{equation}
where H denotes the cross entropy function, $x_{\text{atom}}[\text{mask}]$ denotes the masked atom tokens and $x_{\text{patom}}[\text{mask}]$ denotes the corresponding predicted atom tokens for the masked positions.

In the coordinate denoising task, to increase the challenge of the pertaining task, we introduce Gaussian noise with a standard deviation of 0.2 for all the atom coordinates. Additionally, to enhance broader applicability across downstream applications, we mask atomic features $x_{\text{atomic}}$, bond features $x_{\text{bond}}$, and shortest path distance features $x_{\text{SPD}}$ with a probability of 50\%. Furthermore, we align the conformation of the noised molecule, denoted as $r_{\text{noised\_coor}}$, with that of the raw molecule, denoted as $r_{\text{coor}}$, using the Kabsch algorithm.

In contrast to Uni-Mol, Uni-Mol2 employs the position prediction head to predict the atom coordinates $r_{\text{pcoor}}$ of molecules. 

\begin{equation}
\begin{aligned}
\label{eq: position prediction head}
\Delta_{pos} &= \text{Dis}(r_{\text{noised\_coor}}) \\
Q_{pos} &= \text{FFN}(\text{LN}(x^{N})), \space 
K_{pos} = \text{FFN}(\text{LN}(x^{N})) \\
V_{pos} &= \text{FFN}(\text{LN}(x^{N})),  \space 
B_{pos} = \text{FFN}(\text{LN}(p^{N})) \\
attn_{pos} &= \text{softmax}(Q_{pos}K_{pos}^{T}+B_{pos}) \circ \Delta_{pos} \\
\Delta_{vpos} &= attn_{pos}V_{pos}, \space 
\Delta_{ppos} = \text{FFN}(\Delta_{vpos}) \\
r_{\text{pcoor}} &= r_{\text{noised\_coor}} + \Delta_{ppos}
\end{aligned}
\end{equation}
where $Dis$ denotes element-wise subtraction of positions between different noised atoms $r_{\text{noised\_coor}}$. Specifically, the difference in position between atoms $i$ and $j$ is given by $\Delta_{pos}(i,j) = r_{\text{noised\_coor}, i} -  r_{\text{noised\_coor}, j}$. And $\circ$ denotes Hadamard product. $LN$ denotes layer normalization. $FFN$ denotes a feed-forward network. In practice, we use multi-head attention; for simplicity in writing, we omitted the notation related to heads here.
Once the predicted coordinates $r_{\text{pcoor}}$ are obtained, the predicted pair-distance $r_{\text{pdistance}}$ can be derived by calculating the Euclidean distances between each pair of $r_{\text{pcoor}}$. We integrated coordinate prediction and pair-distance prediction with $\ell_{1}$ loss into Uni-Mol2's optimization process for the coordinate denoising task:

\begin{equation}
\begin{aligned}
\label{eq:coor loss}
\mathcal{L}_{\text{coor}} &= \lVert r_{\text{pcoor}} - r_{\text{coor}}\rVert_{1}, \\
\mathcal{L}_{\text{distance}} &= \lVert r_{\text{pdistance}} - r_{\text{distance}}\rVert_{1}
\end{aligned}
\end{equation}

We eliminated two stabilizing regularization terms from the Uni-Mol model, yielding the final loss of Uni-Mol2:
\begin{equation}
\label{eq:total loss}
\mathcal{L}_{\text{total}} = \mathcal{L}_{\text{atom}} 
 + \mathcal{L}_{\text{coor}} + \mathcal{L}_{\text{distance}} 
\end{equation}

\begin{table*}
    \caption{Architecture of Uni-Mol2 at different scale}
    % \vspace{8pt}
    \centering
    \begin{adjustbox}{max width=\linewidth}
    \begin{tabular}{c|c|c|c|c|c|c|c|c}
        \toprule
            Params     & Layers     & \thead{Embedding \\ dim} & \thead{Attention \\ heads} & \thead{Pair embedding \\ dim} & \thead{Pair hidden \\ dim} & \thead{FFN embedding \\ dim} & \thead{Learning \\ rate}  & \thead{Batch \\ size} \\
        
            \midrule
            42M & 6 & 768 & 48 & 512 & 64& 768& 1e-4& 1024 \\
            
            84M & 12 & 768 & 48 & 512 & 64& 768& 1e-4& 1024 \\
            
            164M & 24 & 768 & 48 & 512 & 64& 768& 1e-4& 1024 \\
            
            310M & 32 & 1024 & 64 & 512 & 64& 1024& 1e-4& 1024 \\
            
            570M & 32 & 1536 & 96 & 512 & 64& 1536& 1e-4& 1024 \\
            
            1.1B & 64 & 1536 & 96 & 512 & 64& 1536& 1e-4& 1024 \\
        \bottomrule
    \end{tabular}
    \label{tab:table scale of model}
    \end{adjustbox}
\end{table*}

\subsection{Hyperparameter and Training Details}
\label{sec:Hyperparameter}
We study the scalability of Uni-Mol with the scale from 42M to 1.1B, and all the parameters for Uni-Mol2 at different scales are listed in Table \ref{tab:table scale of model}. And Uni-Mol2 is trained with AdamW optimzer\cite{kingma2014adam,  loshchilov2017decoupled-adamw}, with the following hyper-parameters: $ \beta{1} = 0.9 $  and $ \beta{2} = 0.99 $ and weight decay $ 1e-4 $. The gradient clip norm is set to 1.0 for training stability. The learning rate scheduler employed is a polynomial decay scheduler during pretraining. Specifically, all models reach its maximum learning rate value $ 1e-4 $ after 100,000 warm-up steps and decay the learning rate of each parameter group using a polynomial function with power 1.0. All the models are trained with mix-precision\cite{micikevicius2017mixed-precision} for training efficiency.

Using the temperature-based sampling method outlined in Equation \ref{sample equation}, we sample 838 million conformations as training samples from the dataset. All models were subsequently trained on these 838 million samples. All these conformations were generated using the ETKGD method \cite{riniker2015better-etkgd} and optimized with the Merck Molecular Force Field (MMFF) \cite{halgren1996merck-mmffs} in RDKit. For models containing parameters ranging from 42M to 310M, we employed 32 NVIDIA A100 GPU cards, while for models with 570M and 1.1B parameters, we utilized 64 NVIDIA A100 GPU cards.

\section{Scaling Laws}

\begin{figure*}
    \centering
    \includegraphics[width=\textwidth]{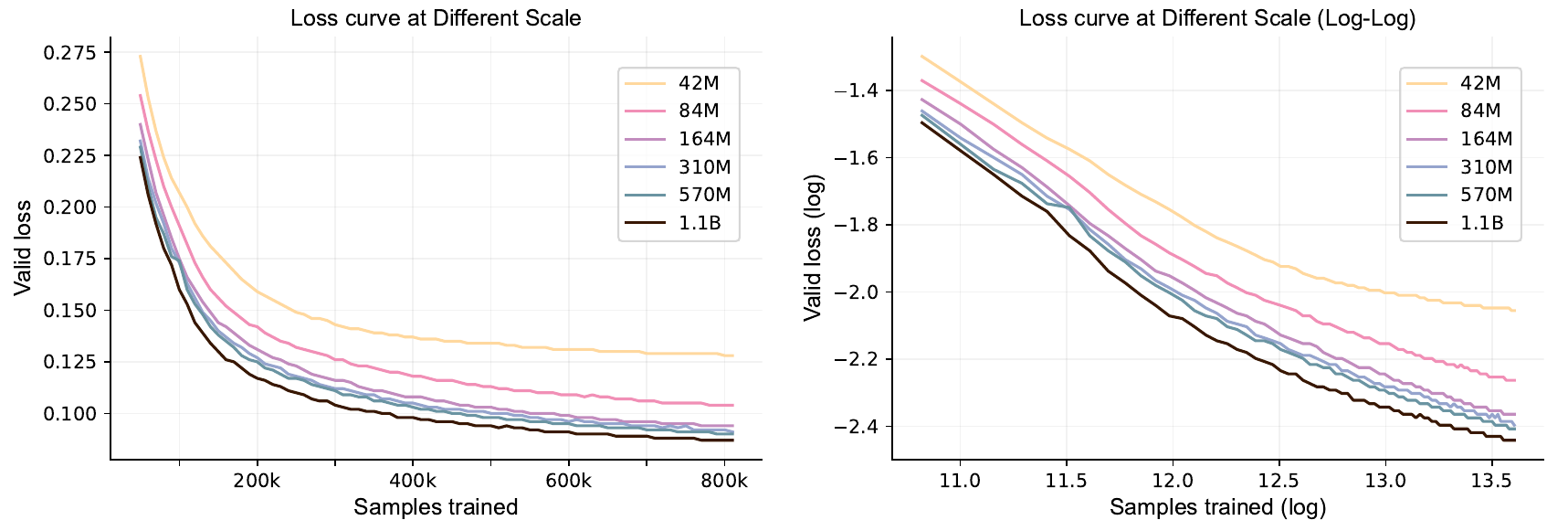}
    \caption{Validation loss curves. Training curves for Uni-Mol2 model from 42M to 1.1B parameters. Models are trained on 0.8B samples. At the convergence stage, the 84M parameters model has a loss of 0.105, and the 1.1B parameters model reaches a loss of 0.087. }
    \label{Uni-Mol2 valid loss curve}
\end{figure*}

Several studies\cite{deepseek-llm, kaplan2020scaling, su2024unraveling} on large language models (LLMs) investigate the power-law connections between model performance, commonly assessed by validation or test loss, and factors such as the number of model parameters, dataset size, and compute budget. Here, we aim to define the power-law of validation loss $\mathcal{L}$ during the model's convergence period. In Figure \ref{Uni-Mol2 valid loss curve}, we present the validation loss of Uni-Mol2 models with parameter counts varying from 42 million to 1.1 billion during the training process. We mainly examine the impact factors of three aspects: data scale $N$, model scale $M$, and compute budget scale $C$. Given that a constant batch size $B$ of 1024 is maintained for Uni-Mol2 across various scales, the number of training steps $S$ is considered as a suitable proxy for $D$, as $D$ can be approximated by the product $BS$.

We initially designed a power term for $M$ and $S$ separately. Additionally, we approximate the computed budget $C$ as $MS$. Notably, we have neglected the intricate relationship between actual computing costs $C$ and $MS$, instead subsuming it into the parameter estimation. Adhering to the design principles of \cite{kaplan2020scaling}, the loss function $\mathcal{L}(M, D)$ should exhibit scale invariance, limit consistency, and analyticity to ensure stability and consistency across varying parameters. As a result, we derived the following empirical power-law relationship:

\begin{equation}
\label{eq: scale formulation}
\mathcal{L}(M, S, C) = \alpha_{m}M^{\beta_{m}} + \alpha_{s}S^{\beta_{s}}  + \alpha_{c}C^{\beta_{c}} 
\end{equation}

We established the relationship based on the validation loss trajectory of Uni-Mol2 across different scales, as detailed in Table \ref{tab:table scale of model}. Specifically, we utilized the validation data from Uni-Mol2 42M, 84M, 164M, and 310M, recording the validation loss every 10,000 training steps. Furthermore, to prevent the performance during the transient period from affecting the parameter estimation, we excluded the loss of information from the first 200,000 training steps. Consequently, we have:

\begin{equation}
\label{eq: scale detail}
\mathcal{L}(M, S, C) = 2.660M^{-1.137} + 1.848S^{-0.225}  + 0.588C^{-1.479} 
\end{equation}

\begin{figure*}
    \centering
    \includegraphics[width=\textwidth]{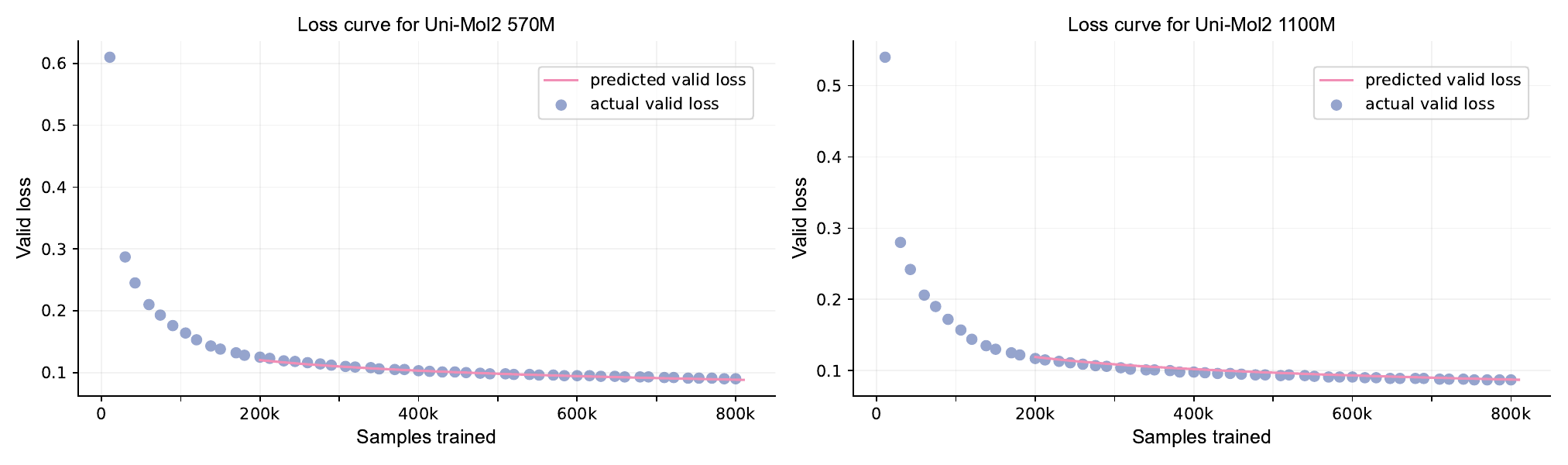}
    \caption{Graph of actual loss and prediction loss across different updates for the 570M (\textbf{left}) and 1.1B (\textbf{right}) models}
    \label{Uni-Mol2 prediction loss curve}
\end{figure*}

As shown in Fig \ref{Uni-Mol2 prediction loss curve}, equation \ref{eq: scale detail} fits the actual validation loss well for Uni-Mol2 570M and Uni-Mol2 1.1B parameters model, particularly when the model's performance reaches convergence. To assess the scaling law's effectiveness, we calculated Relative Mean Absolute Error (RMAE), Mean Square Error (MSE), R-squared, and Pearson Correlation Coefficient by comparing predicted validation loss with actual validation loss over the last 100,000 steps for Uni-Mol2 570M and Uni-Mol2 1.1B on Table \ref{tab:scale performance}. The high Pearson Correlation Coefficient and R-squared we computed indicate a strong linear relationship between our predicted values and the actual data. The RMAE values for Uni-Mol2 570M and Uni-Mol2 1.1B are 0.0169 and 0.0095, respectively, indicating that Equation \ref{eq: scale detail} accurately models the loss curve. Specifically, for the Uni-Mol2 570M at 810,000 steps, the actual validation loss was recorded at 0.09, compared to a predicted loss of 0.088, yielding a predicted validation error of 2.22\%. Meanwhile, for Uni-Mol2 1.1B at the same step, the actual validation loss stood at 0.087, slightly below the forecast of 0.0871, with a prediction error of 0.23\%.

\graphicspath{{}}

% Todo 
\graphicspath{{}}

\begin{table*}
\caption{Metrics about Scaling Law for Uni-Mol2}
% \vspace{8pt}
\centering
\begin{adjustbox}{max width=\linewidth}
\begin{tabular}{c|cccc}
    \toprule
        Model     & RMAE & MSE & R-Squared & \thead{Pearson Correlation \\ Coefficient} \\
    
        \midrule
        
        Uni-Mol2 570M  & 0.0169 &2.450e-6 & 0.92 & 0.85   \\
        
        Uni-Mol2 1.1B  & 0.0095 &8.458e-5 & 0.87 & 0.75 \\
    \bottomrule
\end{tabular}
\label{tab:scale performance}
\end{adjustbox}
\end{table*}

\section{Downstream Experiment}

Upon pretraining with extensive unlabeled datasets using the predefined task, one should acquire a highly accurate molecular representation for fine-tuning downstream tasks. In this section, we conduct experiments on the ability of scaled models on downstream tasks.

\subsection{QM9 Dataset}
We employ QM9 \cite{ramakrishnan2014qm9, wu2018moleculenet} datasets to evaluate the performance of the molecular pretraining model at different scales and compare Uni-Mol2 with representative existing methods. QM9 dataset provides the geometric, energetic, electronic, and thermodynamic properties of the molecule, comprising 134 thousand stable organic molecules with up to nine heavy atoms. Due to QM9 containing several quantum mechanical properties with different quantitative ranges, each property is treated as a separate task. However, the HOMO, LUMO, and HOMO-LUMO GAP, which share similar ranges, are trained together as a single task for simplicity \cite{fang2022geometry-gem}.

\paragraph{Baselines} We evaluate Uni-Mol2 against several baseline models, with a primary emphasis on pretraining baselines. Given that Uni-Mol demonstrates superior performance compared to these baselines in previous work \cite{zhou2023unimol}, our analysis concentrates on the comparison between Uni-Mol and Uni-Mol2, specifically examining the scalability of Uni-Mol2 at various scales. It is noted that we have shifted the dataset partitioning method from scaffold-based partitioning to scaffold similarity-based partitioning, thereby increasing the task difficulty to evaluate the model's performance more comprehensively. The dataset is then divided into training, validation, and test sets in proportions of 80\%, 10\%, and 10\%, respectively. Following previous work \cite{fang2022geometry-gem, zhou2023unimol}, we report the mean and standard deviation by the results of 3 random seeds.

\paragraph{Results} The results are presented comprehensively in Table \ref{tab:qm9 performance}, where the best results are marked in bold. Uni-Mol still outperforms baselines on almost all downstream datasets. Uni-Mol2 outperforms Uni-Mol in four out of the six tasks examined. But as the model parameters increase, Uni-Mol2 demonstrates significantly improved performance, surpassing Uni-Mol across all tasks at the 1.1 billion parameter level, achieving an average 27\% improvement on the QM9 task for all properties. We systematically investigate the scaling of Uni-Mol2 across parameter sizes ranging from 84 million to 1.1 billion. Except for the $C_{v}$ property prediction task, the results for other properties progressively improve as the model size increases, consistent with the patterns observed in the model's validation performance. This indicates that enlarging the model consistently enhances downstream performance. However, for properties such as HOMO, LUMO, HOMO-LUMO GAP, and ZPVE, the results converge as the model size increases. This convergence suggests that further increases no longer influence the performance ceiling for these tasks in model size.

\begin{table*}
\caption{Mean absolute error(MAE, $\downarrow$) results on QM9 Dataset}
% \vspace{8pt}
\centering
% \begin{small}
\begin{adjustbox}{max width=\linewidth}
\begin{tabular}{c|ccccccc}
    \toprule
        Model     & HOMO / LUMO / GAP & alpha & $C_{v}$ & mu & $R^{2}$ & ZPVE \\
    
        \midrule
        $GROVER_{base}$ & 0.0079 (3e-04) & 2.365 (0.302) & 1.103 (0.339) & 0.618 (0.002) & 113.01 (4.206) & 0.0035(3e-04) \\

        $GROVER_{large}$ & 0.0083 (6e-04) & 2.240 (0.385) & 0.853 (0.186) & 0.623 (0.006) & 85.85 (6.816)  & 0.00381(5e-04) \\
        
        GEM & 0.0067(4e-05) & 0.589(0.0042) & 0.237(0.0137) & 0.444(0.0015) & 25.67(0.743) & 0.0011(2e-05) \\
        
        Uni-Mol & 0.0043(2e-05) & 0.363(0.009) & 0.183(0.002) & 0.155(0.0015) & 4.805(0.055) & 0.0011(3e-05)  \\
        
        Uni-Mol2 84M & 0.0038(5e-05) & 0.376(0.027) & 0.178(0.012) & 0.105(0.0009) & 4.968(0.235) & 0.0010(1e-04) \\
        
        Uni-Mol2 164M & 0.0036(1e-05) & 0.325(0.004) & 0.157(0.017) & 0.093(0.0006) & 4.935(0.189) & 0.0005(1e-05) \\
        
        Uni-Mol2 310M & 0.0036(1e-05) & 0.315(0.003) & \textbf{0.143(0.002)} & 0.092(0.0013) & 4.672(0.245) & 0.0005(1e-05) \\
        
        Uni-Mol2 570M & 0.0036(2e-05) & 0.315(0.004) & 0.147(0.0007)  & 0.089(0.0015) & 4.523(0.080) & 0.0005(3e-05) \\
        
        Uni-Mol2 1.1B & \textbf{0.0035(1e-05)} & \textbf{0.305(0.003)} & 0.144(0.002) & \textbf{0.089(0.0004)} & \textbf{4.265(0.067)} & \textbf{0.0005(8e-05)} \\
    \bottomrule
\end{tabular}
\end{adjustbox}
% \end{small}
\label{tab:qm9 performance}
\end{table*}

\subsection{COMPAS-1D Dataset}
Due to the QM9 dataset only providing the conformation, some molecules failed to generate the atom and bond feature correctly. Therefore, fine-tuning Uni-Mol2 on the existing QM9 dataset to evaluate its effectiveness with bond and edge features presents a non-trivial challenge. To further validate the performance and generalization capabilities of the Uni-Mol2 pretraining model, we  
utilized COMPAS-1D from COMPAS project \cite{wahab2022compas}. COMPAS-1D offers essential computational properties crucial for comprehending the behaviour of polycyclic aromatic hydrocarbons and other organic molecules across various chemical and physical processes. Modeling the relationships of these properties has significant implications for the field of organic photoelectric materials.

We still follow the QM9 scaffold similarity-based partition and split it by a ratio of 8:1:1 into the train, validation, and test sets. Table \ref{tab:photoelectric performance} presents the predictive capabilities of Uni-Mol2 regarding photoelectric quantum properties. The model with $\star$ suffix indicates that they incorporate atom and bond features.  The results indicate that Uni-Mol2 excels in all tasks except for $\text{aEA}$ property prediction task. Additionally, consistent with findings from the QM9 dataset, Uni-Mol2 demonstrates superior performance across all tasks as the model scales up. The results also show that under the same parameter scale, models incorporating atom and bond features outperform those without these features. Uni-Mol2 1B achieves 4\% improvement over Uni-Mol, while Uni-Mol2 1B with atom and bond feature achieves 14\% improvement over Uni-Mol. This suggests that, in certain scenarios, these features consistently provide a significant advantage.

\begin{table*}
\caption{
\centering
Mean absolute error(MAE, $\downarrow$) results on COMPAS-1D Dataset. }
% \vspace{8pt}
\centering
\begin{adjustbox}{max width=\linewidth}
\begin{tabular}{c|cccc}
    \toprule
        Model     & aEA & aIP & dispersion & Dipmom Debye \\ 
    
        \midrule
        Uni-Mol & 0.0099(2e-05) & 0.0083(9e-05) & 0.0092(6e-04) & 0.0198(2e-04) \\

        Uni-Mol2 84M & 0.0104(2e-05) & 0.0081(3e-05) & 0.0092(5e-04) & 0.0196(1e-04)   \\

        Uni-Mol2 1.1B & \textbf{0.0103(4e-04)} & \textbf{0.008(1e-05)} & \textbf{0.0081(1e-04)} & \textbf{0.0186(3e-04)} \\

        \midrule
        
        Uni-Mol2 84M $\star$ & 0.0104(4e-04) & 0.0077(5e-05) & 0.0085(1e-04) & 0.0173(6e-04)  \\
        
        Uni-Mol2 1.1B $\star$  & \textbf{0.0093(4e-05)} & \textbf{0.0074(9e-05)} & \textbf{0.0067(2e-04)}  & \textbf{0.0170(2e-04)} \\
    \bottomrule
\end{tabular}
\label{tab:photoelectric performance}
\end{adjustbox}
\end{table*}

\subsection{The Performance on Limited QM9 Dataset}

In numerous fields like bio-medicine, acquiring extensive well-annotated molecular data is often expensive and time-consuming. Typically, these datasets include only a limited quantity of data\cite{butler2018machine, li2022improving}. To evaluate the performance of Uni-Mol2 with restricted data availability, we conducted sampling on the QM9 dataset. We sampled the training set by stratifying it according to the quantile binning of the HOMO-LUMO GAP label from the QM9 test set and then created subsets named train50, train100, and train200 by sampling at 50\%, 100\%, and 200\% of the test set size, respectively.

We enhanced Uni-Mol2 from 84M to 1.1B parameters using train50, train100, and train200 datasets to predict HOMO, LUMO, and GAP properties on the QM9 test dataset. As illustrated in Table \ref{tab:qm9 limited performance}, two conclusions emerge from the MAE for predicting HOMO, LUMO, and HOMO-LUMO GAP on the QM9 test set. First, the model's performance, indicated by a decreasing MAE, progressively improves as the training dataset expands. This is evident from comparing the MAE values between the train50 and train200 rows across different scales of the Uni-Mol2 models. For example, the Uni-Mol2 84M model shows a reduction in MAE from 0.0062 to 0.0046, marking a 25.8\% decrease as the dataset grows from 50 to 200 instances. Secondly, in situations where training data is scarce, the larger Uni-Mol2 models demonstrate enhanced predictive capabilities. This is evidenced by the fact that the Uni-Mol2 1.1B parameters model, which has the largest parameters, consistently records the lowest MAE scores for all sizes of training sets. This is especially apparent in the train50 scenario, where it achieves an MAE of 0.0056, marking the best performance among the models discussed. These results highlight the advantages of enlarging both the training dataset and the model scale to improve predictive accuracy in downstream finetuning tasks with Uni-Mol2.

\begin{table*}
\caption{Mean absolute error(MAE, $\downarrow$) about HOMO-LUMO GAP on QM9  Dataset}
\centering
\begin{adjustbox}{max width=\linewidth}
\begin{tabular}{c|cccc}
    \toprule
        Model     & \fontsize{8}{8} train50 & train100 & train200  \\
    
        \midrule

        Uni-Mol2 84M & 0.0062(8.1e-05) & 0.0053(1.0e-06) & 0.0046(1.0e-06)  \\
        
        Uni-Mol2 164M & 0.0058(3.7e-05) & 0.0050(1.4e-05) & 0.0044(6.9e-05)  \\
        
        Uni-Mol2 310M & 0.0056(4.7e-05) & 0.0049(0.4e-06) & 0.0044(4.0e-05)  \\
        
        Uni-Mol2 570M & 0.0057(4.2e-05) & 0.0048(1.8e-05) & 0.0044(8.1e-06)  \\
        
        Uni-Mol2 1.1B & \textbf{0.0056(1.8e-05)} & \textbf{0.0048(3.5e-05)} & \textbf{0.0043(4.7e-05)}  \\
    \bottomrule
\end{tabular}
\label{tab:qm9 limited performance}
\end{adjustbox}
\end{table*}

\section{Conclusion}

In this paper, to fully investigate the scaling law in the molecular pretraining field, we construct a diverse dataset of molecular structures spanning 884 million instances and present a novel molecular pretraining model Uni-Mol2. We successfully scale the model size to 1.1 billion parameters from 84 million parameters and characterize the power-law relationship between validation loss and model size, dataset size, and computational resources. By empowering the power-law relationship of Uni-Mol2, it can shed light on the performance of the larger model. Our largest 1.1B parameters model also outperforms the existing methods.
% todo 总结这部分需要润色一下

The scaling law paves the way for exploring larger models to achieve higher performance. We hope that our work can open avenues for further exploration of the foundational molecular pretraining model. While larger models yield substantial benefits, there are still several potential future directions. Firstly, beyond property prediction tasks, it is also worthwhile to explore whether the representation can be effectively utilized to enhance generative tasks. Secondly, even though the Uni-Mol2 has shown excellent results in several domains by increasing model capacity, it remains to be explored whether the advantages of scaling are beneficial for a broader range of tasks. Thirdly, the current mainstream large language models (LLMs) are predominantly based on a decode-only architecture. It is worth investigating whether there are more elegant decode-only architectures for molecular pre-training models.

{\small
\printbibliography
}

%%%%%%%%%%%%%%%%%%%%%%%%%%%%%%%%%%%%%%%%%%%%%%%%%%%%%%%%%%%%

\newpage

\appendix

\section{Implementation Details}

\subsection{Dataset Description}

\paragraph{QM9 Dataset} The QM9 dataset \cite{ramakrishnan2014qm9} is a significant resource in the field of quantum chemistry, offering a single equilibrium conformation and 12 labels that include geometric, energetic, electronic, and thermodynamic properties. For the purpose of performance evaluation, we select the following properties: HOMO, LUMO, gap, alpha, $C_{v}$, mu, $R^{2}$, and ZPVE. The details of the properties are as follows:

\begin{itemize}

\item \textbf{HOMO}  The HOMO (Highest Occupied Molecular Orbital) is the highest energy molecular orbital that is occupied by electrons in a molecule. 

\item \textbf{LUMO} The LUMO (Lowest Unoccupied Molecular Orbital) is the lowest energy molecular orbital that is not occupied by electrons.

\item \textbf{gap} The gap, often referred to as the HOMO-LUMO gap, is the energy difference between the HOMO and LUMO. It is a measure of the energy required to excite an electron from the HOMO to the LUMO. 

\item \textbf{ZPVE} ZPVE (Zero-Point Vibrational Energy) is the energy associated with the vibrational motion of atoms in a molecule at absolute zero temperature.

\item \textbf{$\alpha$} The $\alpha$ value represents the static polarizability of a molecule. 

\item \textbf{$C_{v}$} The $C_{v}$ (Heat Capacity at Constant Volume) is the amount of heat needed to raise the temperature of a given amount of substance by one degree Celsius at constant volume.

\item \textbf{$\mu$} The $\mu$ (Dipole Moment) is the measure of the molecule's permanent electric dipole moment. 

\item \textbf{$r^{2}$} The $r^{2}$ (Electronic Spatial Extent) is defined as the expectation value of the square of the electronic distance from the nucleus.

\end{itemize}

\paragraph{COMPAS-1D Dataset} The COMPAS-1D dataset is a part of the COMPAS Project, which is an acronym for the computational Database of Polycyclic Aromatic Systems. The dataset is specifically focused on data-condensed poly-benzenoid hydrocarbons, which are a type of polycyclic aromatic hydrocarbons (PAHs) with a unique structure where the benzene rings are connected edge-to-edge. The COMPAS-1D \cite{wahab2022compas} contains 8,678 molecules and offers essential computational properties crucial for comprehending the behavior of polycyclic aromatic hydrocarbons and other organic molecules across various chemical and physical processes. The details of the properties used in the downstream tasks are as follows:

\begin{itemize}
    \item \textbf{aEA}  aEA (Adiabatic Electron Affinity) measures the tendency of a molecule to gain an electron.
    \item \textbf{aIp} aIP (Adiabatic Ionization Potential) measures the energy required for a molecule to lose an electron.
    \item \textbf{Dispersion} Dispersion describes weak inter-molecular forces important for understanding molecular interactions.
    \item \textbf{Dipmom Debye}  Dipmom in Debye indicates the polarity of a molecule, affecting its interactions and solubility.
\end{itemize}

\subsection{Atom and Bond Feature for Molecules}

The molecular feature used in Uni-Mol2 contains two parts: 1) Atom and bond features, we use RDkit to generate these atom and bond features as input of Uni-Mol2. The detailed features are listed in Table \ref{atom feature} and Table \ref{bond feature}. 2) Shortest path $\text{SPD}_{i,j} $. We employ the Floyd-Warshall algorithm\cite{hougardy2010floyd} to calculate the shortest distances between each pair of connected atoms.

\begin{table*}[h]
  \caption{Atom features}
  \label{small}
  \centering
   \small
   \begin{adjustbox}{max width=\linewidth}
  \begin{tabular}{c|c|c}
    \toprule
    \textbf{features} & \textbf{size} & \textbf{description} \\ 
    \midrule

    atom type & 119 & type of atoms including C, N, O, etc, by atomic number \\
    chirality & 6 & type of chirality like Tetrahedral chirality \\
    degree & 11 & the degree of an atom in molecule \\
    formal charge & 11 & integer electronic charge assigned to atom \\ 
    number of H & 9 & number of bonded hydrogen atoms \\ 
    number of radical electrons & 5 & number of radical electrons \\
    hybridization & 5 & SP, SP2, SP3, SP3D, SP3D2 \\ 
    aromaticity & 1 & whether an atom is part of an aromatic system \\ 
    in ring  & 1 & whether an atom is within a ring structure \\

    \bottomrule
  \end{tabular}
  \label{atom feature}
  \end{adjustbox}
\end{table*}

\begin{table*}[h]
  \caption{Bond features}
  \label{small}
  \centering
   \small
   \begin{adjustbox}{max width=\linewidth}
  \begin{tabular}{c|c|c}
    \toprule
    \textbf{features} & \textbf{size} & \textbf{description} \\ 
    \midrule

    bond type & 4 & SINGLE, DOUBLE, TRIPLE, AROMATIC \\
    stereo & 6 &  NONE, Z, E, CIS, TRANS, ANY, \\
    conjugated & 1 & whether the bond is conjugated \\

    \bottomrule
  \end{tabular}
  \label{bond feature}
  \end{adjustbox}
\end{table*}

\subsection{Hyperparameter Settings}

In line with previous methods, we employ grid search to find the optimal hyper-parameters for tasks within the QM9 and COMPAS-1D datasets. The specific hyper-parameters are detailed in Table \ref{finetune hyper-parameter}. In all experiments, we select the checkpoint with the lowest validation loss and report the corresponding test set results based on that checkpoint. For the COMPAS-1D dataset, experiments were conducted using a single A100 GPU, whereas for the QM9 dataset, the experiments were run on eight A100 GPUs.

\begin{table*}[h]
  \caption{Hyper-paramters for fine-tuning on QM9 and COMPAS-1D Dataset }
  \label{small}
  \centering
   \small
   \begin{adjustbox}{max width=\linewidth}
  \begin{tabular}{l|c}
    \toprule
    Hyperparameter & Value or description\\
    \midrule
    Learning rate      &  [4e-5, 6e-5, 1e-4, 2e-4, 3e-4, 4e-4] \\
    Batch size  &  [32, 64, 128]  \\    
    Epochs      &  [40, 60, 80, 100, 200, 300] \\
    Pooler dropout &  [0.0, 0.1] \\
    Warmup ratio   & [0.0, 0.06, 0.1] \\
    % GPUs & \thead{8 GPUs for QM9 \\ 1 GPU for COMPAS-1D} \\
    \bottomrule
  \end{tabular}
  \label{finetune hyper-parameter}
  \end{adjustbox}
\end{table*}

\subsection{Evaluation Metrics}
Diverse evaluation metrics can better help us understand and evaluate the effectiveness of our model. In this section, we introduce the evaluation metrics used in this study. Given $n$ samples, where $y_{i}$ is the actual value and $\hat{y}_{i}$ is the predicted value.

Mean Absolute Error (MAE) calculates the average of the absolute differences between predicted and actual values in regression tasks, treating errors of different scales equally.

\begin{equation} 
\label{eq:MAE} 
\text{MAE} = \frac{1}{n} \sum_{i=1}^{n} |\hat{y}_i - y_i| 
\end{equation}

 Relative Mean Absolute Error (RMAE) measures the average absolute prediction error relative to the actual values, providing a dimensionless indication of model accuracy. By normalizing with the actual values, it removes the effect of the data scale, making it possible to compare data with different scales. 
\begin{equation}
\label{eq:RMAE}
\text{RMAE} = \frac{1}{n} \sum_{i=1}^{n} \frac{|\hat{y}_i - y_i|}{|y_i|}
\end{equation}

Mean Square Error (MSE) calculates the average of the squared differences between predicted and actual values, heavily penalizing larger errors.

\begin{equation}
\label{eq:MSE}
\text{MSE} = \frac{\sum_{i=1}^{n} (\hat{y}_i - y_i)^2}{n}
\end{equation}

R-squared measures the proportion of variance in the dependent variable that can be predicted by the independent variables, highlighting the goodness of fit for a regression model. A higher R-squared indicates that the independent variables explain a significant portion of the variance in the dependent variable, while a lower R-squared indicates that the model explains less.
\begin{equation}
\label{eq:R-squared}
\text{R-squared}= 1 - \frac{\sum_{i=1}^{n} (y_i - \hat{y}_i)^2}{\sum_{i=1}^{n} (y_i - \bar{y}_i)^2}
\end{equation}

The Pearson Correlation Coefficient (r) measures the linear correlation between two variables, ranging from -1 to 1. Larger absolute values signify a stronger linear relationship between the two variables, while values near 0 indicate a weak or non-existent linear relationship.

\begin{equation}
\label{eq:Pearson Correlation Coefficient}
r = \frac{\sum_{i=1}^{n} (x_i - \bar{x})(y_i - \bar{y})}{\sqrt{\sum_{i=1}^{n} (x_i - \bar{x})^2 \sum_{i=1}^{n} (y_i - \bar{y})^2}}
\end{equation}

\section{Infrastructures}

We utilize an efficient distributed PyTorch framework called Uni-Core \cite{Uni-Core}, specifically designed for swiftly developing high-performance PyTorch models \cite{ansel2024pytorch}, particularly those based on Transformer architectures\cite{vaswani2017attention-transformer}. Given the variability in molecule lengths, padding inputs to match the maximum molecular length is necessary during training. Consequently, the batch size for model training is influenced by the longest molecule in each batch. However, since molecule lengths follow a long-tail distribution (with the majority falling within a specific range), we employ dynamic batching techniques to enhance GPU utilization. By adjusting batch sizes according to the maximum lengths of different batches, we can significantly boost GPU utilization with minimal effort. 

The time consumption of reading data from distributed storage is often overlooked. We employ a singular, dedicated process on each computational node to asynchronously replicate the training dataset of each epoch onto the host machine. This strategy effectively mitigates time overheads, thereby obscuring the duration spent on data reading from distributed storage. To resume the corruption due to the infra and other factors effectively, we save model weight and optimizer state for every 1k step asynchronously. This means we will lose 1k step training resources in the worst case of hardware instability or loss spike during training. Meanwhile, any checkpoints exceeding the most recent ten files will be deleted to avoid consuming too much storage space.

\section{Limitations}
The major limitation of our study pertains to the absence of an exploration of the optimal batch size and learning rate. Our investigation primarily focuses on analyzing and delineating the power-law relationships among validation loss, model size, dataset size, and computational resources. The predictive accuracy of performance aligns well with the scaling curve, indicating that the current optimal learning rate and batch size approximate the near-optimal values. However, existing research suggests a progressive increase in the optimal batch size with augmented computing resources, while the optimal learning rate tends to decrease gradually. It is necessary to note that as we further increase the model's parameters, the final optimal values for learning rate and batch size may fall outside the currently identified range. Consequently, investigating the scaling law for optimal batch size and learning rate is also paramount.

% lies pertains to its interpretability. Although our model demonstrates enhanced effectiveness and efficiency, it falls short in terms of interpretability compared to traditional docking methods.These conventional approaches offer visualizations that elucidate the binding mechanism between a pocket anda molecule, providing clear explanations.

%%%%%%%%%%%%%%%%%%%%%%%%%%%%%%%%%%%%%%%%%%%%%%%%%%%%%%%%%%%%
\end{document}